\documentclass[conference, letterpaper]{IEEEtran}
\usepackage[noadjust]{cite}  
\usepackage{lineno,hyperref}
\usepackage{amssymb}
\usepackage[cmex10]{amsmath}
\usepackage{float}
\usepackage{verbatimbox}
\usepackage{fancyvrb}
\usepackage{bm}
\usepackage[bbgreekl]{mathbbol}
\usepackage{amsfonts}
\usepackage{booktabs}
\usepackage{romannum}
\usepackage{tablefootnote}
\usepackage{algorithm}
\usepackage[noend]{algpseudocode}
\DeclareSymbolFontAlphabet{\mathbb}{AMSb}
\DeclareSymbolFontAlphabet{\mathbbl}{bbold}
\usepackage{booktabs} 
\usepackage[utf8]{inputenc}
\usepackage[caption=false,font=footnotesize]{subfig}
\usepackage{array}
\usepackage{multirow}
\usepackage{blkarray}

\DeclareMathOperator*{\argmaxA}{arg\,max} 

\ifCLASSINFOpdf
\else
\fi
\ifCLASSINFOpdf
   \usepackage[pdftex]{graphicx}
\else
\fi

\usepackage{color}
\usepackage{fancyhdr}
\usepackage[caption=false,font=footnotesize]{subfig}

\renewcommand{\thispagestyle}[2]{}

\fancypagestyle{plain}{
        \fancyhead{}
        \fancyhead[C]{first page center header}
        \fancyfoot{}
        \fancyfoot[C]{first page center footer}
}
\begin{document}

%
\title{Reinforcement Learning for Fair Dynamic Pricing}


\author{\IEEEauthorblockN{Roberto Maestre, Juan Duque, Alberto Rubio, Juan Arevalo}
\IEEEauthorblockA{BBVA Data and Analytics\\
Av. de Burgos, 16D, 28036 \\ Madrid, Spain \\
\{roberto.maestre, juanramon.duque, alberto.rubio.munoz, juanmaria.arevalo\}@bbvadata.com}}



%


\maketitle

\begin{abstract}
Unfair pricing policies have been shown to be one of the most negative  perceptions customers can have concerning pricing, and may result in long-term losses for a company. Despite the fact that dynamic pricing models help companies maximize revenue,  fairness and equality should be taken into account in order to avoid unfair price differences between groups of customers. This paper shows how to solve dynamic pricing by using  Reinforcement Learning (RL) techniques so that prices are maximized while keeping a balance between revenue and fairness. We demonstrate that RL provides two main features to support fairness in dynamic pricing: on the one hand, RL is able to learn from recent experience, adapting the pricing policy to complex market environments; on the other hand, it provides a trade-off between short and long-term objectives, hence integrating fairness into the model's core. Considering these two features, we propose the application of RL for revenue optimization, with the additional integration of fairness as part of the learning procedure by using Jain's index as a metric. Results in a simulated environment show a significant improvement in fairness while at the same time maintaining optimisation of revenue.
\end{abstract}



%
\IEEEpeerreviewmaketitle

\section{Introduction}
Determining the right price of a product or service for a particular customer is a necessary, yet complex endeavour; it requires knowledge of the customer's willingness to pay, estimation of future demands, ability to adjust strategies to competition pricing \cite{deksnyteDP}, etc. Dynamic pricing \cite{narahari2005dynamic,Boer:DP} represents a promising solution for this challenge due to its intrinsic adjustment to  customer expectations. Indeed, with the advent and establishment of digital channels, unique opportunities for the application of dynamic pricing are arising, thus enhancing research in the field \cite{JanetAdamy, RePEc:sym}. Equality is a critical aspect of dynamic pricing as it influences customers'  perceptions of fairness \cite{garbarino2003dynamic, lee2011perceived, xia2004price, weisstein2013effects, Dynamic_pricing_fairness, Odlyzko:2003}. As long as perceptions of fairness are used by people as a heuristic for trust, inequality may lead to a destruction of that trust. For instance, large price differences between groups of customers, or a disagreement between the distribution of cost and profit are known to affect the relationship between customers and sellers \cite{kimesretrospective}. Should this relationship be damaged, it could eventually generate substantial financial losses for a company in the medium- and long-term \cite{kahneman1986fairness}. Although the  definition of fairness depends on the domain context and, consequently, has  a diffuse definition~\cite{finkel2001notfair}, we propose clear design principles for fairness in dynamic pricing with a  focus on elements such as group or individual schemes (in which policy decisions are applied), and equity or definitions of equality. This clear design allows us to specify a well-defined metric and to include it in the overall model in order to ensure fairness.

Different approaches have been proposed to address the problem of maximizing revenue \cite{narahari2005dynamic}. Among these is a promising technique consisting of optimizing pricing policies with Reinforcement Learning (RL) applied to different market scenarios (uni or multi-agent) \cite{KUTSCHINSKI20032207, kononen2006dynamic, GuptaRL1168003, MenonRLDimamic6873894}. Despite the extensive application of RL models to dynamic pricing, concepts such as equality and fairness are rarely incorporated into the learned policies. Moreover, the \emph{black--box} nature of Machine Learning models~\cite{fairMLauth, MLOpacity,bostromSuperintell, Cerquitelli:2017:TDM:3110841} means that it is of utmost importance to ensure a good trade-off between revenue (goal for companies) and equality (principle of fairness and a goal for customers) \cite{Mikians}. There are also ethical issues in artificial intelligence such as algorithm bias, in which if the input data (or market environment itself) reflects unfair biases of the broader society  \cite{tolga-biasGender,bath55288,hardtPS16,heeBiasFaces}, the output model can potentially  capture such biases perpetuating unfair policies. These biases can have a significant  impact on dynamic pricing outcomes, as the efficiency of policies (revenue) has to be balanced against the need to achieve equality treatment of costumers \cite{weiss2001online}. 

Many metrics have been proposed in order to measure fairness from the \emph{resource allocation distribution} point of view~\cite{TianLan-Resources}. Such a perspective  has been widely studied \cite{arianpoo2016network, francis2012techniques, zhang2010tcp} in order to avoid unfair resource allocation for nodes in wireless networks. In the present paper we apply concepts from resource allocation distribution to dynamic pricing so as to  interpret fairness as \emph{similarly distributed prices among groups of customers} (equality between groups of customers). Identification of different groups of customers is required for an effective price discrimination. In each group, certain assumptions about price sensitivity have to be made; groups could be defined, for instance, by customer's income level, gender, location, communication channel, etc. Our learning procedure provides homogeneous prices among such groups, but at the same time, takes into account price sensitives within each group to maximize revenue. As we will show, maintaining a balance of price distribution among groups of customers will boost fairness perception and thus trust, as the price for different offers is chosen by considering equality.

The present study is organized as follows. Section \ref{secc:background} provides some useful definitions from the fairness design principles, the resource allocation distribution literature, and reviews the main concepts of RL. The experimental methodology is then presented in section \ref{secc:experiment}, where synthetic groups of customers are introduced, and RL is applied to the dynamic pricing problem. Next, we show our experimental results in section \ref{secc:results}. Finally, we draw some conclusions in section \ref{secc:conclusions}.

\section{Background}\label{secc:background}
In this section we will start by first reviewing main concepts of Reinforcement Learning (RL), more specifically Q-Learning as variation of RL. We also define the approximation of Q-values by means of Neural Networks (NN) minimizing a loss function. Thereafter we will spell out the design principles for fairness. Finally, we will  introduce the main Jain's index properties, which will allow us to measure fairness in dynamic pricing policies.

\subsection{Reinforcement Learning concepts}
Reinforcement Learning is an area of machine learning that studies how an agent takes actions in an environment to achieve a given goal \cite{Sutton:1998}. RL differs from supervised learning as the agent learns by trial and error while interacting with the environment, as opposed to learning with labeled data. As such, RL is best suited for learning within sequential decision problems.

Q-learning is a variant of RL introduced by Watkins \cite{Watkins1992}, whereby the heuristics of the model are directly related to the rewards provided by the environment in each iteration. Q-Learning can be applied to dynamic pricing as each action will modify the state of the environment (related to fairness), thus providing a reward (the bid itself). Equation (\ref{eq:q_learning}) represents the generic way in which Q-learning is expressed:
\begin{equation}
\label{eq:q_learning}
\begin{split}
Q(s, a) \leftarrow (1-\alpha)Q(s, a) + \alpha \Big( r' + \gamma \argmaxA_{a^*} Q(s',a^{*}) \Big\rvert s,a \Big).
\end{split}
\end{equation}
Here, $\alpha$ is the learning rate, $\gamma$ the discount factor and $r'$ is the reward obtained after performing action $a$ in state $s$. Thus, the goal of RL is to find a policy $\pi_*: s \rightarrow a$ that maximizes the expected discounted utility with some exploration methods. We use $\epsilon$-greedy strategy that follows a random action with $\epsilon$ probability and action with $1-\epsilon$ probability.

Q-Learning in its simplest form uses a table (Q-table) to represent state-action values, i.e. Q-Values. The problem, however, becomes intractable  as the number of states and actions increases \cite{KUTSCHINSKI20032207, Dini12combiningq-learning}. In general terms, the Q-value function in equation (\ref{eq:q_learning}) can be estimated by function approximation, i.e. $Q(s,a) \approx Q(s,a;\theta)$ \cite{van2007reinforcement}. 
Since NN act as universal function approximators \cite{Hornik:1989}, they provide an excellent framework for Q-value estimation.
Indeed, recent advances in NN applied to Q-learning have shown them to greatly outperform traditional RL methods \cite{Riedmiller-approximation, Lin-RL-NN, Dini12combiningq-learning, VolodymyrAtari}. 
The process of learning Q-value estimations is as follows: at each iteration step $i$, approximated Q-values $Q(s,a;\theta_i)$ are trained by Stochastic Gradient Descent, by minimizing the loss $L(\theta_i)$:
\begin{equation}
\label{eq:q_network}
\begin{split} L(\theta_i) =\left[ \Big( y - Q(s,a; \theta_i) \Big)^2 \right],
\end{split}
\end{equation}
where $y = \left[  r' + \gamma \underset{a^*}{\argmaxA} Q(s',a^{*}; \theta_{i-1}) \Big\rvert s,a \right]$.

Reinforcement learning algorithms constitute a suitable method for learning  pricing policies, whenever the expected revenue for taking a pricing action is unknown in the absence of complete information of the environment \cite{RANA2015426}.

\subsection{Fairness Design Principles}
\label{secc:fairness_design}

We say that a policy in dynamic pricing is fair when the policy provides \emph{similarly distributed prices among groups of customers}. From this definition, we can establish the next principles: A) $equality$, which drives homogeneous prices, and B) $groups$ as atomic $entities$ for applying prices, rather than individuals. Taken these two principles into account, we further propose to use Jain's Index \cite{jain1999throughput} as fairness metric because: I) it works on multiple groups, II) it provides an indicator of how homogeneous is the policy as a clear indicator of how much the policy is fair in percentage terms (from 0 to 100\% fair). Other metrics for measuring equality and fairness suited to be applied to dynamic pricing,  such as entropy, min-max ratio, etc.  are out of the scope of this paper. A complete and comprehensive study on several fairness metrics and their properties can be found in the reference list \cite{TianLan-Resources}.

\subsection{Metrics for Fairness: Jain's Index}
\label{secc:metrics_fairness}
Let $\mathcal{C}$ be a portfolio of customers. Each customer $c_i$ belongs to one group $g_\alpha \in \mathcal{G}$, where $\mathcal{G}$ is a collection of groups covering $\mathcal{C}$. We use $\bar{g}_\alpha$ to represent the average price allocated in group $g_\alpha$. 

The Jain's index is defined as
\begin{equation}
J = \frac{\left(\sum_{g\in\mathcal{G}} \bar{g}\right)^2}{|\mathcal{G}|\sum_{g\in\mathcal{G}} \bar{g}^2}.\nonumber
\end{equation}
It provides a measurement of how fair the average price allocation is in $\mathcal{G}$ (here, $|\mathcal{G}|$ is the number of groups in $\mathcal{G}$). Jain's index poses two convenient characteristics when used to define the heuristics in a RL model: I) values are continuous between $[0,1]$ and II) the index applies to any number of groups. First feature I) fits perfectly with our definitions of states in Section \ref{secc:states} while the second II) is valuable when several groups of customers are defined. 

\begin{figure}[H]
\label{fig:jains_index}
\begin{center}
\includegraphics[width=2.5in] {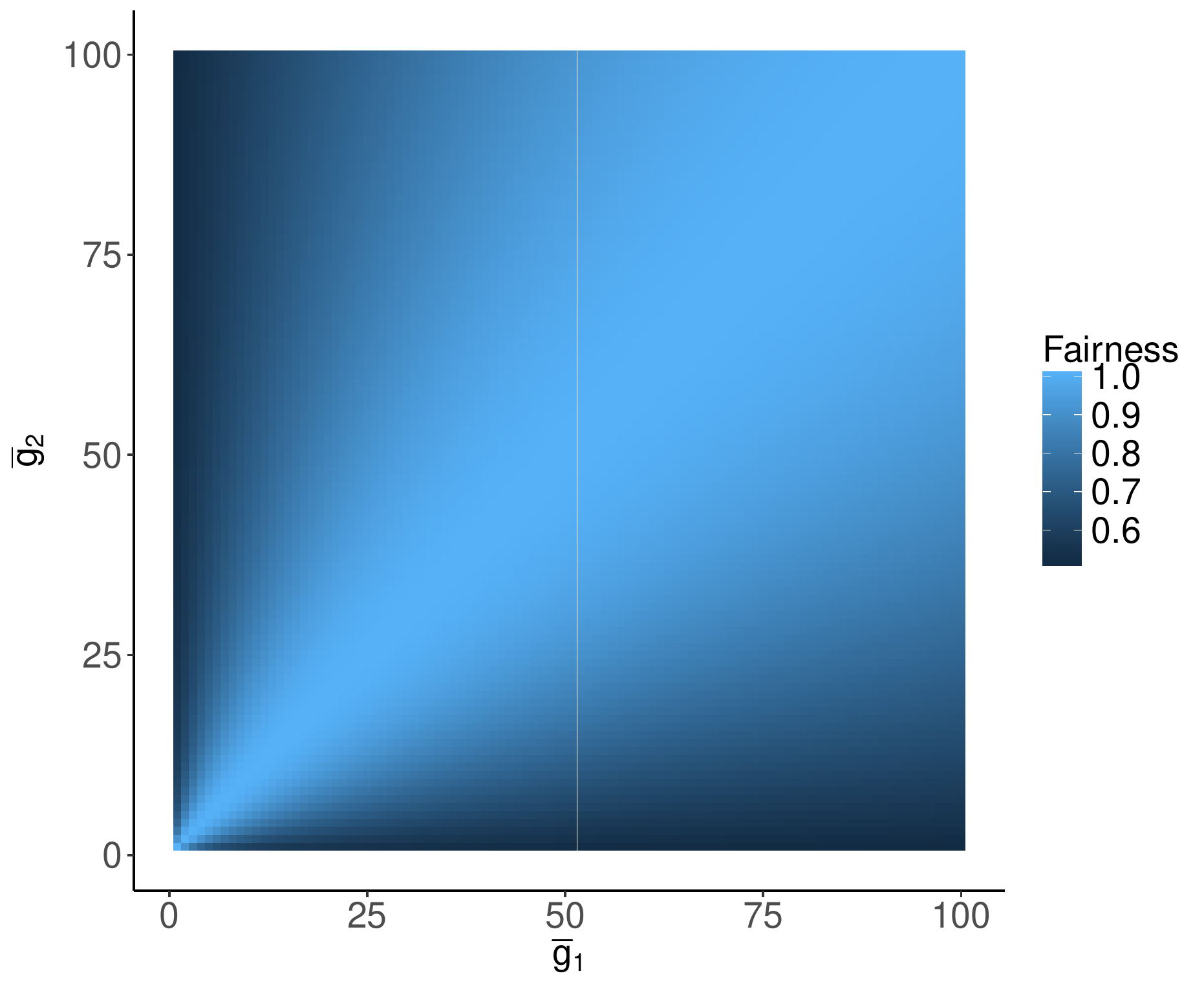}
\caption{Jain's index for two groups of customers. Please, notice that the index equals here the percentage of people for which a pricing policy is fair.}
\label{fig:jains_function}
\end{center}
\end{figure}

As we mentioned before, Jain's index offers an excellent way to know how fair our price allocation is among groups in $\mathcal{G}$. For instance, if there are only two groups and either $\bar{g}_1 \ll \bar{g}_2$ or $\bar{g}_2 \ll \bar{g}_1$, Jain's index is $0.5$, reflecting an unfair situation. Therefore, the model should learn to increase prices when possible in group $g_{1}$ and/or decrease the prices in $g_{2}$. This two-group case is represented in figure \ref{fig:jains_function}. As observed, values on the diagonal are 1.0, because both groups have the same average price, i.e.: $\bar{g}_1=\bar{g}_2$. Thus, the policy is fair with an homogeneous price between groups. However, when $\bar{g}_1=50, \bar{g}_2=0$ or $\bar{g}_1=0, \bar{g}_2=50$ the policy is fair only with the $50 \%$ of the groups (heterogeneous case). Thus, by means of this metric, we can constrain the differences between groups and provide fairer prices.

\section{Methodology}
\label{secc:experiment}

We define four synthetic groups $\mathcal{G} = \{g_1,g_2,g_3,g_4\}$ to experiment with fairness and revenue. We simulate customers $c \in \mathcal{C}$ belonging to any group in $\mathcal{G}$. Each customer responds to each price bid with the following logistic function $\phi \in [0,1]$:
\begin{equation}
\begin{split}
\label{eq:elasticity}
\phi(a,g \mid a \in \mathcal{A}, g \in \mathcal{G}) = \big[1+e^{-\left(b+w\cdot a
\right)}\big]^{-1}
\end{split}
\end{equation}\normalsize
where $a$ is the action chosen by the agent (the bid), and $b_g$ and $w_g$ are parameters defining the sensitivity of each group, see Table \ref{tab:logistic_parameters}.

\begin{table}[htbp]
   \centering
   \caption{ Weights for the logistic function $\phi$ }
   \label{tab:logistic_parameters}
   \begin{tabular}{|c|cc|}
     \hline
     Group & $b$ & $w$\\
     \hline
     $1$ & 
     $18.229$ & 
     $-2.369$ \\
     
     $2$ & 
     $4.4757$ & 
     $-1.1526$ \\
     
     $3$ & 
     $-1.09195$ & 
     $0.34000$ \\
     
     $4$ & 
     $0$ & 
     $0$ \\
     
     \hline
\end{tabular}
\end{table}

\begin{figure}[H]
\begin{center}
\includegraphics[width=2.5in] {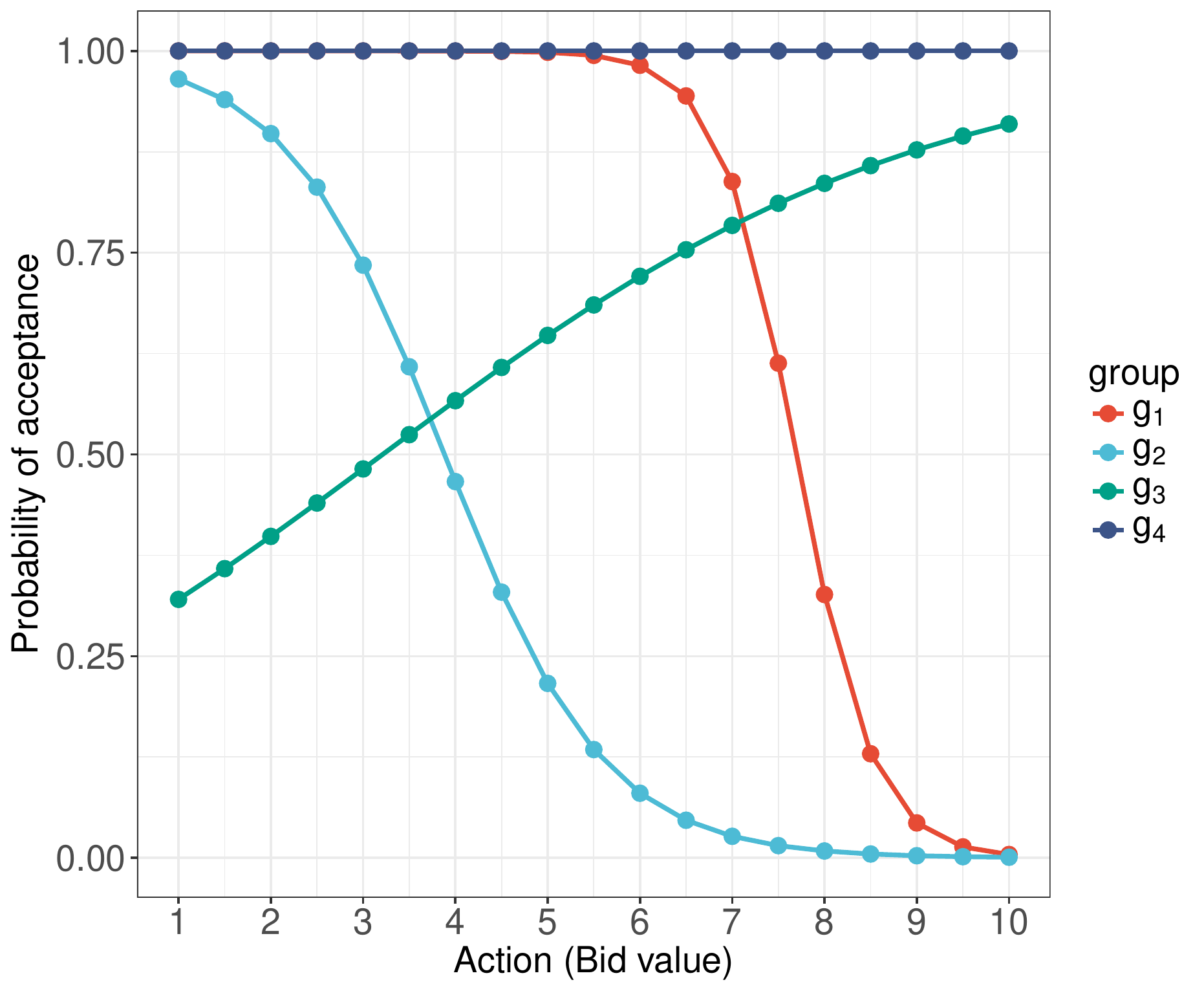}
\caption{Prices acceptance probabilities for the $4$ groups defined in Equation (\ref{eq:elasticity}) and Table \ref{tab:logistic_parameters}.}
\label{fig:elasticites}
\end{center}
\end{figure}

This simple environment provides a scenario in which four different  behaviours are simulated. Customers belonging to $g_1$ will accept much higher prices than customers in $g_2$. On the other hand, customers from $g_3$ exhibit an inverse behaviour: the larger the price, the higher the acceptance probability. This group can be seen as customers whose perception of the product quality increases with price. Finally, $g_4$ customers are not sensible to any change in the price. Figure \ref{fig:elasticites} represents the probability of acceptance of a given price for each group (agent action or bid value).

In order to achieve both the maximum revenue and fairness, the model should learn to increase the price within each group (maximizing the revenue), while reducing the differences between groups when possible, so that fairness is also maximized. Please note that maximizing revenue entails finding the maximum price for each group in which the probability of acceptance is the highest \cite{phillips2005pricing}.

The statistics in the experiments are computed by running $350$ epochs. In each epoch $1000$ bids are simulated with a $100$ customers randomly selected from $\mathcal{C}$. In order to achieve a good balance between exploration and exploitation, for every epoch $t$ we decrease $\epsilon$ according to the next ratio $\epsilon = 1/e^{\frac{t}{20}}$ (starting by $\epsilon = 1$).

The following subsections define the main elements of RL proposed in the present paper for a fair dynamic pricing. To this end, we define a state that allows us to identify both the customer group, $g \in \mathcal{G}$, and the average allocated price, $\bar{g}$; this way, both revenue and fairness can be maximized. As will be shown, revenue maximization is included in the reward through the value of each bid (see equation (\ref{eq:reward})).

\subsection{States}
\label{secc:states}
We define the state as $\mathcal{S} = \{ s_{c}, s_{f}\}$, where $s_{c}$ represents the customer group, and $s_{f}$ represents the global fairness. We propose to code $s_{c}$ and $s_{f}$ as independent one-hot vectors. For the latter, $s_f$, we define a partition $\mathcal{I}$ of mesh $p$ in the interval $[0,1]$, $\mathcal{I}\equiv 0< p<2p< ... < 1 - p< 1$, so that there is a $k$ such that $\mathcal{I}_k\le s_f\le \mathcal{I}_{k+1}\, \forall s_f$. Given this partition, we define the one-hot encoding of $s_f$ as a vector with a $1$ at position $k$, and zeros otherwise. Finally, $\mathcal{S}$ is a vector resulting from the concatenation of $s_{c}$ and $s_{f}$ one-hot encoding, and has the final dimension $n=|\mathcal{G}|+1/p+1$. As an example, if we define a partition with mesh $p=1/3$, one state with customer $c_{\alpha} \in g_2$ and a global Jain's index with value $0.89$, the vector  $\mathcal{S}$ will be represented as follows:
$$ \mathcal{S}_{\alpha} =
    \begin{blockarray}{cccccccc}
    \multicolumn{4}{c}{$\mathcal{G}$} & \multicolumn{3}{c}{$J$} \\
    \multicolumn{4}{c}{$\overbrace{\rule{3.0cm}{0pt}}$} &
    \multicolumn{3}{c}{$\overbrace{\rule{3.0cm}{0pt}}$} \\
      g_1 & g_2 & g_3 & g_4 & [0,\frac{1}{3}) & [\frac{1}{3},\frac{2}{3}) & [\frac{2}{3}, 1]   \\
      \downarrow &\downarrow & \downarrow &\downarrow &\downarrow &\downarrow &\downarrow \\
    \begin{block}{[cccccccc]}
      0&1&0&0&0&0&1  \\
    \end{block}
    \end{blockarray} 
$$

\subsection{Actions}
We define the action space, $\mathcal{A}$, as a partition of mesh $q$ in the interval $[\min, \max]$, i.e. 
\begin{equation*}
\mathcal{A}\equiv \min < \min+q< \min+2q< ...< \max-q< \max.
\end{equation*}
The action space has dimension $m=1/q+1$. Please note that as long as $\mathcal{A}$ is constrained between $\{\min, \max\}$ prices, the model will never bid excessively unfair prices --i.e., a non-sense price. 

Extensive experimentation show that best values of $p,q$ for discretizing the state and action vectors lie at around $0.01$. With this selection we provide a good balance between expansion and shrinkage over the state/action space. On the one hand, a fine grid mesh (large $p$ and $q$)  make certain over--the--space defined functions (such as customer sensitivity in equation (\ref{eq:elasticity})) more expressive. On the other hand, small discretizations negatively affect learning convergences. Thus, the selection of parameters $p$ and $q$ is critical.

\subsection{Q-value approximation}
In accordance with our definitions of state and action, we propose a simple lineal approximation of the policy:
\begin{equation}
\label{eq:tensor_relu}
\begin{split}
T_0(w_0,b_0): & \mathbb{R}^n \rightarrow \mathbb{R}^m \\
        & s \rightarrow a =  w_0 \cdot X+b_0, 
\end{split}
\end{equation}
where $w_0\in\mathbb{R}^{m\times n}$ and $b_0\in\mathbb{R}^m$ are learnable parameters. 
Thus, we are approximating Q-values with a linear, one-hidden layer network. Depending on the problem complexity, the NN can be expanded with more layers and different activation functions. With this compact method only two vectors (weights and bias for every layer) are needed to represent all Q-values.

\subsection{Reward}
Reward definition in dynamic pricing modeling with RL is generally challenging, as it represents the heuristics of the model itself. In the present paper, a multi-objective heuristic is required to account for both revenue and fairness optimization. Thus, hyper-parameters are needed to balance and be established as specific goals for each objective. For instance, one goal might be learning a policy with a fairness value equal to 0.85 while maximizing the bidding price, or viceversa. Bearing this in mind, we define the reward as:
\begin{equation}
\begin{split}
\label{eq:reward}
r = \beta_p \exp\left(-\cfrac{ ( p -  p_t )^2}{\sigma_p}\right) +\beta_f \exp\left(-\cfrac{ ( f -  f_t )^2 }{\sigma_f}\right).
\end{split}
\end{equation}
The first term in equation (\ref{eq:reward}) stands for price optimization, while the second intends to balance it with fairness. 
Here $\beta_p, \sigma_p, p_t$ and  $\beta_f, \sigma_f, f_t$ are hyper-parameters that balance and establish a specific goal for $p$ and $f$, respectively. Variable $p \in \mathbb{R} [0,1]$ relates to the price given in the $i^{th}$ bidding, with $p_t$ the target price. Similarly, $f \in \mathbb{R} [0,1]$ is the fairness index, and $f_t$ the desired (target) fairness. In order to model the reward, we have chosen a normal distribution of prices and fairness as it exhibits  two convenient properties \cite{Matignon-ImprovingRL}: it points in the direction of the target price $p_t$ and target fairness $f_t$; it also provides a uniform decay for states distant from the main goal. The following figure \ref{fig:reward_function} shows several examples of hyper-parameter choices. NB: The reward is scaled between $[0,1]$.

\begin{figure}[htb]
\label{fig:reward_function}
\begin{center}
\includegraphics[height=2.5in,width=2.5in] {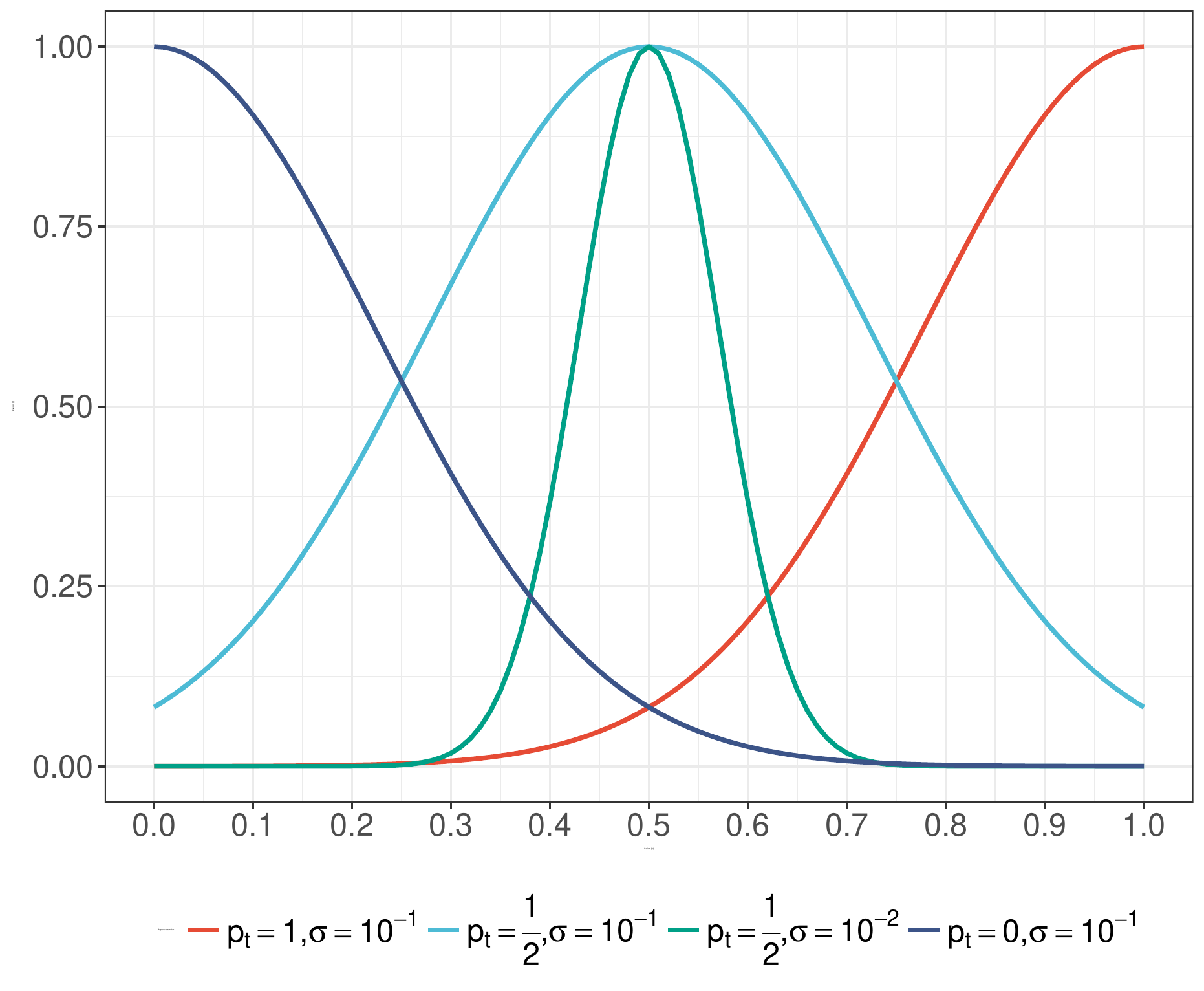}
\caption{Gaussian reward function}
\end{center}
\end{figure}

In a real environment, customers will accept or reject each bid, i.e, the bid will be the value itself or $0$. Thus, we define $p$ as follows:
\begin{equation}
\begin{split}
\label{eq:customer_response}
p =
\left\{
	\begin{array}{ll}
		\nu(a) & \mbox{if } \mathcal{B}\big(\phi (a,g)\big) = 0 \\
		a & \mbox{if } \mathcal{B}\big(\phi(a,g)\big) = 1
	\end{array}
\right.
\end{split}
\end{equation}
Here, $\mathcal{B}$ is the binomial distribution, and the function $\phi(a,g \mid a \in \mathcal{A}, g \in \mathcal{G})$ is given by equation (\ref{eq:elasticity}). The $\nu(a)$ function represents the penalization when a bid is rejected. It could be modeled with a zero value, a constant negative value to penalize the rejection, or even a value proportional to the bid itself. However, in dynamic pricing it is important to keep the rejection rate low because the expected revenue of a policy is calculated as the ratio between the average bid value and the rejection ratio. Hence, it is convenient to take $\nu=$constant. From an industrial perspective, we use a large penalization ($\nu = -0.5$) in order to avoid customer rejection.

\begin{figure}[htb]
\label{fig:jains_index_rotated}
\begin{center}
\includegraphics[width=2.5in] {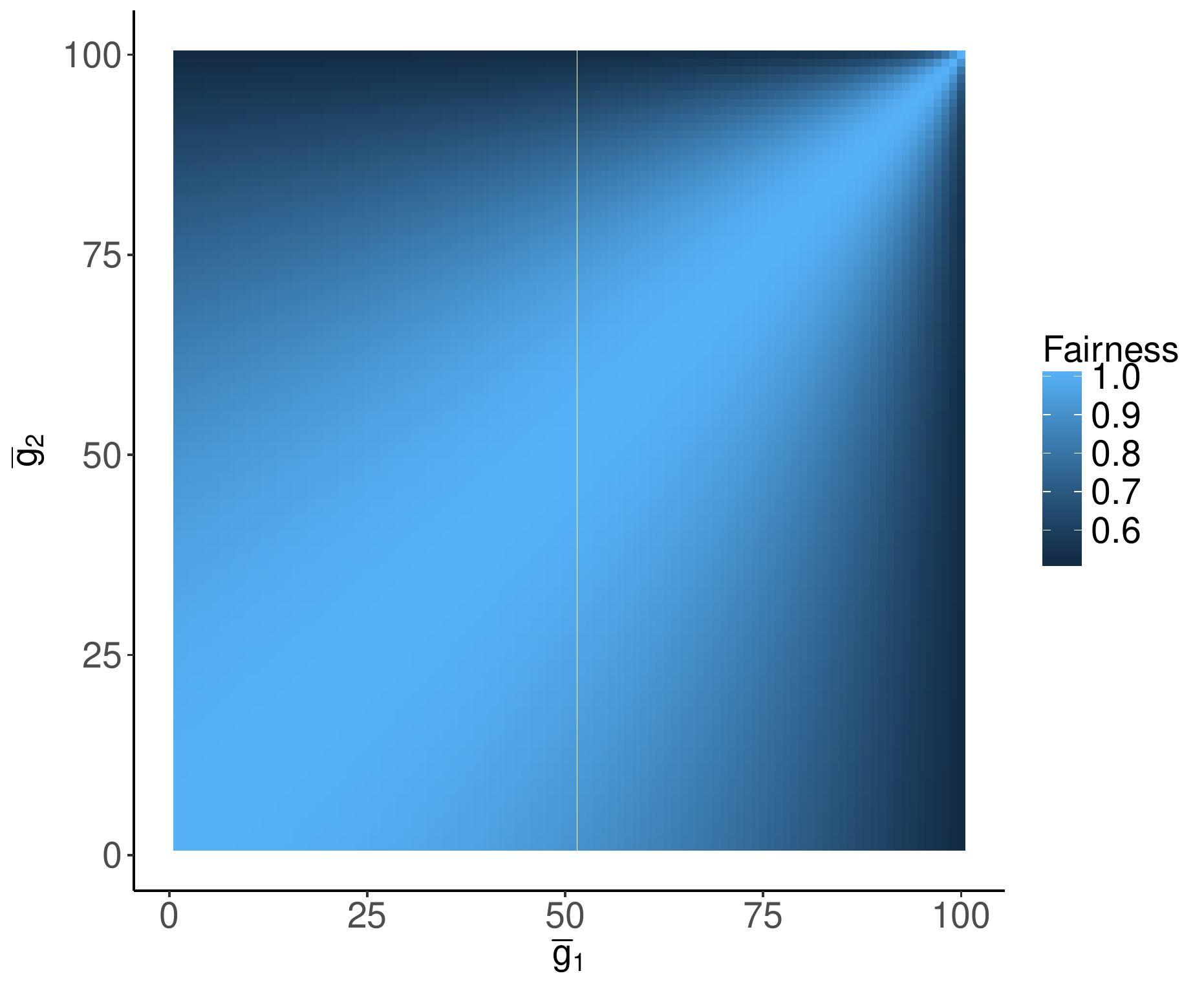}
\caption{Rotated Jain's index}
\label{fig:jains_function_rotated}
\end{center}
\end{figure}

Next, we define the variable $f$, that accounts for fairness optimization in equation (\ref{eq:reward}). From a resource allocation point of view, given a fixed number of groups of customers ($|\mathcal{G}|$ fixed), a large and homogeneous value of the resources ($\bar{g}$ large) will increment Jain's index, and thus fairness. However, in the dynamic pricing context, increasing the average bid price will certainly not be perceived as fair. Therefore, we propose to distribute low and homogeneous prices among groups ($\bar{g}$ low), in contrast to increasing prices where possible to maximize revenue. These ideas can be expressed mathematically as follows:
\begin{equation}
\begin{split}
\label{eq:jain_index}
f = \cfrac{ \Big( \sum_{g \in G} \text{max}\{\mathcal{A}\} - \bar{g} \Big)^2 }{|\mathcal{G}| \sum_{g \in G} \Big( \text{max}\{\mathcal{A}\} - \bar{g} \Big)^2} \in [0,1],
\end{split}
\end{equation}
which can be interpreted as a rotated Jain's index after comparing figures \ref{fig:jains_index} and \ref{fig:jains_function_rotated}. Nevertheless, our transformation keeps the main properties of Jain's index invariant (i.e. population size, scale and metric independence, boundedness or continuity). Despite the simplicity of this approach, this rotation effectively allows for the balance between revenue and fairness.

Finally, we should point out that the selection of reward's hyper-parameters in equation (\ref{eq:reward}), along with the discretization of the state/action space, must be carred out with care as it dramatically impacts the learning process.


\subsection{Reinforcement Learning with NN}
\label{secc:training}

The following pseudo-code defines a standard way for training a NN to approximate Q-values. As illustrated in algorithm \ref{alg:algorithm}, the agent takes each bid $a$ to evaluate the fairness using averages prices in $\mathcal{G}$. The fairness value is calculated with the latest information given by $\mathcal{G}$. With both the bid ($p$) and fairness information (f), the reward $r$ is obtained. Finally, a gradient descent step is executed to reduce the loss produced between the current and the target Q-value, see equation ($\ref{eq:q_network}$). In the experiments, we use the ADAM gradient descend algorithm \cite{KingmaB14} with a learning rate equal to $0.01$.

\begin{algorithm}
\caption{Q-learning with NN approximation}\label{alg:algorithm}
\begin{algorithmic}[1]
\State {$\text{Initialize Neural Network with Random weights}$}
\For{$e=1, E$}
\Comment{$E$ total epochs}
\State {$\forall \bar{g} \in \mathcal{G}, \bar{g} = 0$} \Comment{Initialize  revenue distribution}
\For{$i=1, I$}
\Comment{$I$ total iterations}
\State{
$a_i =
\left\{
	\begin{array}{ll}
		\argmaxA_{a^*} Q(s_i,a^{*}; \theta_{i}) & \mbox{w.p. } 1-\epsilon \\
		\text{Random action} & \mbox{w.p. }\epsilon 
	\end{array}
\right.$
}
\State {bid $a_i$ on the environment} \Comment{One one customer}
\State {Update $\mathcal{G}$}\label{lst:line:independent}
\State {Calculate fairness $f_{i} \text{ based on eq. }(\ref{eq:jain_index})$}  \Comment{On $\mathcal{G}$}
\State {Get reward $r_i \text{ based on eq. }(\ref{eq:reward})$}
\State {Get state $s^{'} = s_{i+1}$}
\State{
$
y_i =
\left\{
	\begin{array}{ll}
		\nu & \mbox{if } $r$_i = 0 \\
		$r$_i + \gamma \argmaxA_{a^{*}} Q(s^{'},a^{*};\theta_{i-1})  & \mbox{o/w }
	\end{array}
\right.
$
}

\State {$\Big( y_i - Q(s,a; \theta_i) \Big)^2$} \Comment{Gradient descent step}
\EndFor
\EndFor
\end{algorithmic}
\end{algorithm}

\section{Results and discussion}\label{secc:results}

The evaluation of the agent's learning procedure is challenging, as the binomial distribution in equation ($\ref{eq:customer_response}$)  introduces statistical fluctuations that lead to changes in the states/actions distribution. In order to alleviate these instabilities, we use the average cumulative bid ($\overline{F_p}$) and average fairness ($\overline{f}$). Although the bid $p$ can take negative values in equation (\ref{eq:customer_response}), we constrain $p \in [0, +\infty)$ for a better representation of the actual revenue. Regarding the reward $r$ in equation (\ref{eq:reward}), we scale the cumulative reward for each experiment between $[0,1]$. This normalization is applied for a better comparison, because in each experiment different weights are used ($\beta$).

We conduct a number of experiments in order to understand the effect of the different competing objectives in equation~(\ref{eq:reward}) (\emph{i.e.}, revenue and fairness).
Experiments $\Romannum{1}$ and $\Romannum{2}$ are designed to test both objectives separately (revenue and fairness). Experiments $\Romannum{3}$ and $\Romannum{4}$ to find a policy with a desired target fairness. Experiment $\Romannum{5}$ is proposed as a null model which has no learning mechanism, thus all actions are randomly selected ($\epsilon = 1.0$) with zero reward value. The following table provides the main results by comparing the average cumulative bid, the averaged fairness, the reject ratio and the expected cumulative revenue for the previous defined experiments. The expected revenue $\overline{r}$ is calculated as result of the product between $\overline{F_p}$ and rejects.

\setlength{\tabcolsep}{4pt}
\renewcommand{\arraystretch}{1.3}

\begin{table}[htbp]
   \centering
   \caption{Results for the proposed experiments}
   \label{tab:fairness_revenue}
   \begin{tabular}{|c|cccc|cccc|}
   \hline
     \multicolumn{1}{|c|}{Exp.} & 
     \multicolumn{4}{c}{Hyperparameters} &
     \multicolumn{4}{|c|}{Results} \\
     \hline
      & $\beta_p$ & $\beta_f$ & $p_t$ & $f_t$ & $\overline{F_p}$  & $\overline{f}$  & Rejects & $\overline{r}$ \\
     \hline
     $\Romannum{1}$ & 1 & 0 & 1 & - & $\textbf{2761.3}$ & 0.62 & 0.10 & $\textbf{2485.17}$ \\
     $\Romannum{2}$ & 0 & 1 & - & 1 &  1326.2 & $\textbf{0.99}$ & 0.25 & 994.65 \\
     \hline
     $\Romannum{3}$ & 1 & 1 & 1 & 0.90 &  2282.1 & 0.88 & 0.11 & 2031.21 \\
      $\Romannum{4}$ & 1 & 1 & 1 & 0.75 &  2527.4 & 0.76 & 0.11 & 2249.36\\
     \hline
     $\Romannum{5}\tablefootnote{$\epsilon$ = 1.0 \text{.In null model all actions are randomly selected}.}$ & 0 & 0 & - & - & 1708.9 & 0.97 & $\textbf{0.31}$ &  1179.12\\
     \hline
\end{tabular}
\end{table}

 For every experiment, table \ref{tab:segments_distributions} shows the average bid allocated in each group. This table provides valuable information related to the fairness achieved for each policy. Related to the fairness principles design defined in section \ref{secc:fairness_design}, we can observe that the more target fairness given by $f_t$ the more homogeneous prices among groups are. Thus, the policy incorporate the proposed fairness principles definition.

\setlength{\tabcolsep}{6pt}

\begin{table}[htbp]
   \centering
   \caption{Average bid on segments $g_{\alpha} \in \mathcal{G}$}
   \label{tab:segments_distributions}
   \begin{tabular}{|c|cccc|}
     \hline
     Exp. & $\overline{g}_1$ & $\overline{g}_2$ & $\overline{g}_3$ & $\overline{g}_4$ \\
     \hline
     $\Romannum{1}$ & 
     $5.8$ & 
     $2.0$ & 
     $9.5$ & 
     $9.8$ \\
     
     $\Romannum{2}$ & 
     $2.7$ & 
     $2.9$ & 
     $2.6$ & 
     $2.7$ \\
     
     $\Romannum{3}$ & 
     $5.6$ & 
     $2.0$ & 
     $6.6$ & 
     $6.8$ \\
     
     $\Romannum{4}$ & 
     $5.5$ & 
     $2.1$ & 
     $8.5$ & 
     $6.6$ \\
     
     \hline
     $\Romannum{5}$ & 
     $5.4$ & 
     $5.5$ & 
     $5.4$ & 
     $5.5$ \\
     
     \hline
\end{tabular}
\end{table}

\begin{figure}[h]
\begin{center}
\includegraphics[width=2.5in] {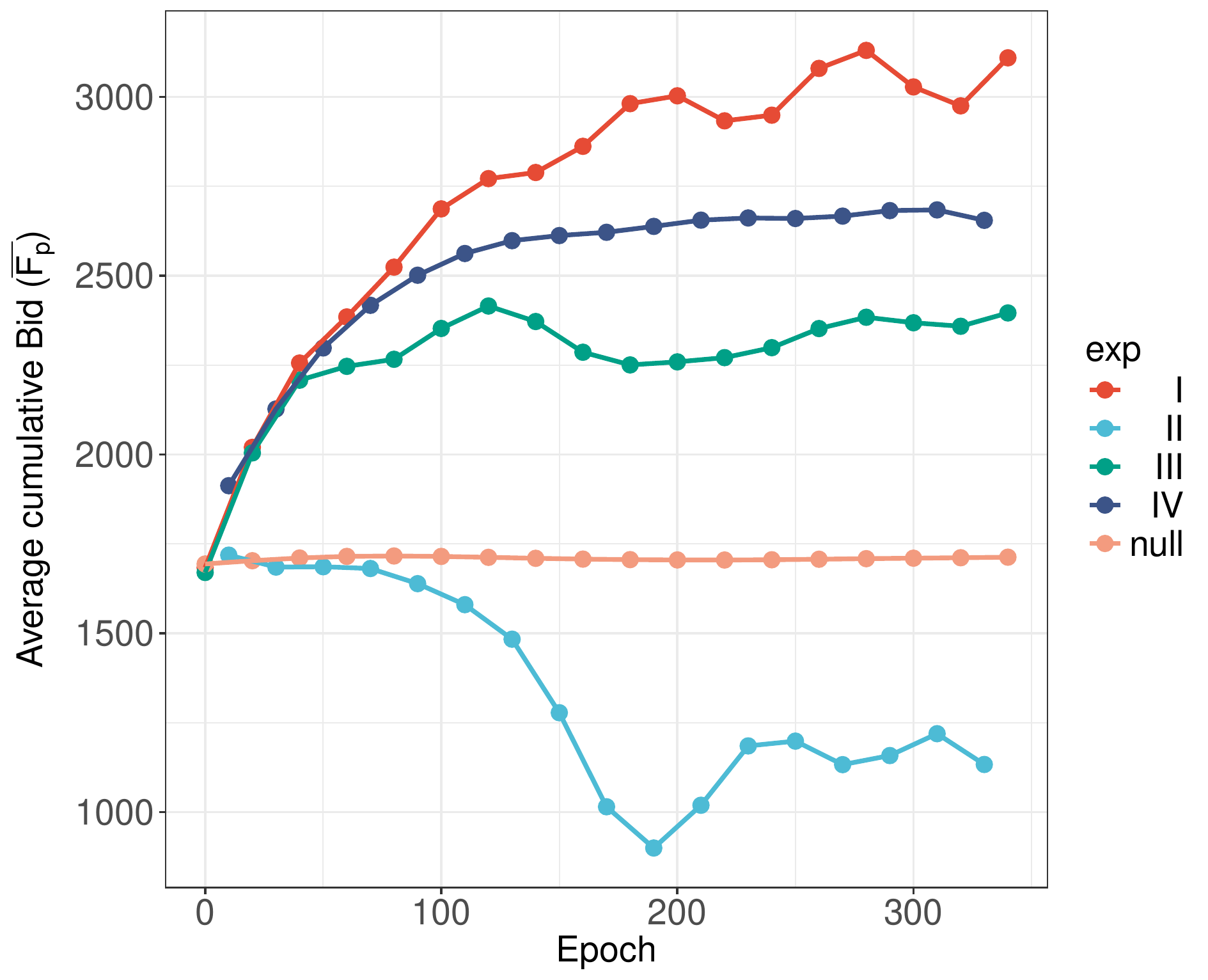}
\caption{Averaged cumulative bid. In exp. $\Romannum{1}$ the model learns to maximize revenue in contrast to exp. $\Romannum{2}$ in which learns to be fair. Exp. $\Romannum{3}$ and $\Romannum{4}$ are proposed to learn a model balancing revenue and fairness with a specific targets. In the exp. $\Romannum{5}$ all actions are taken randomly}
\label{fig:experiment_1_reward}
\end{center}
\end{figure}

\begin{figure}[h]
\begin{center}
\includegraphics[width=2.5in] {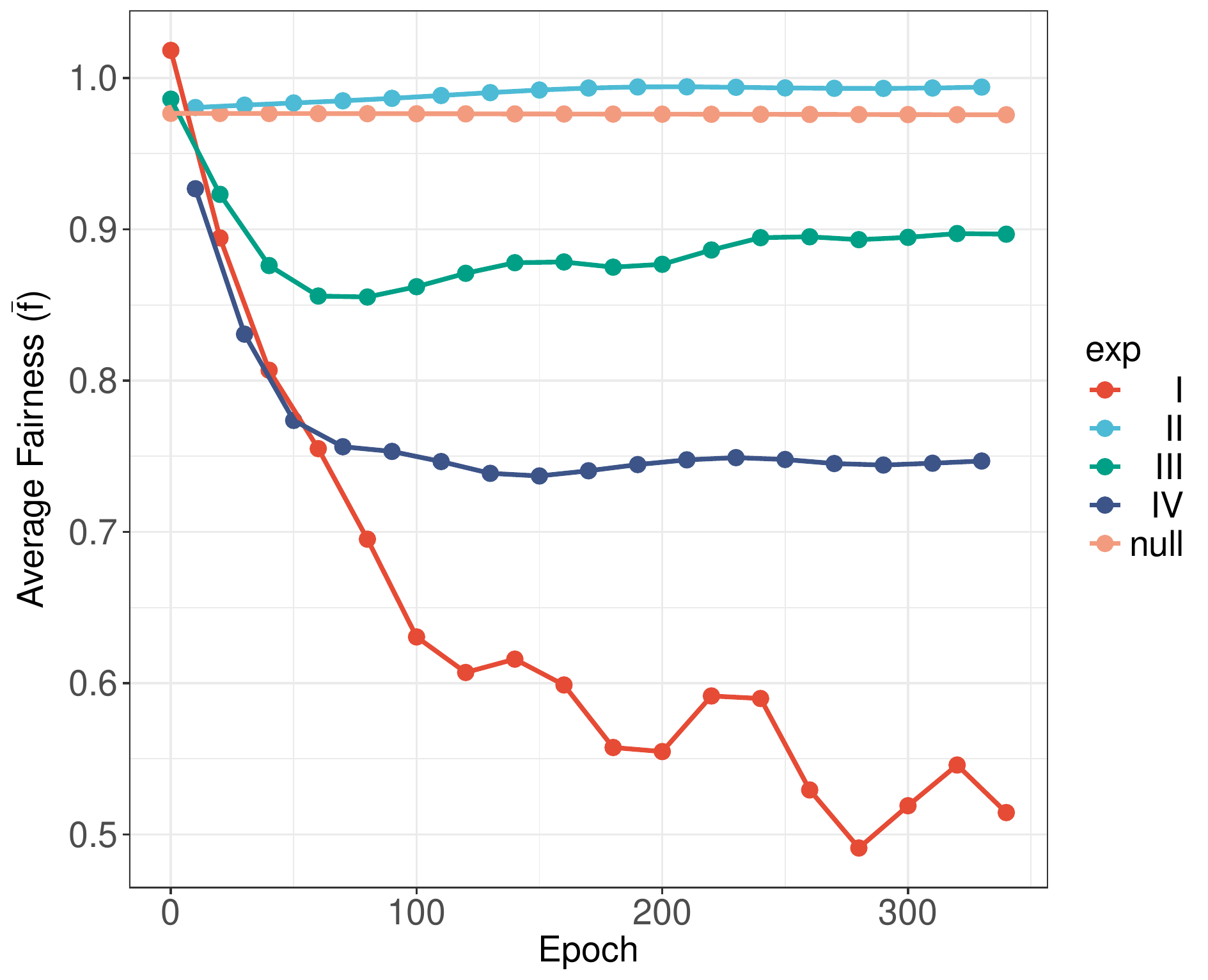}
\caption{Averaged fairness (see description of experiments in figure \ref{fig:experiment_1_reward})}
\label{fig:experiment_1_fairnes}
\end{center}
\end{figure}
	
\begin{figure}[h]
\begin{center}
\includegraphics[width=2.5in] {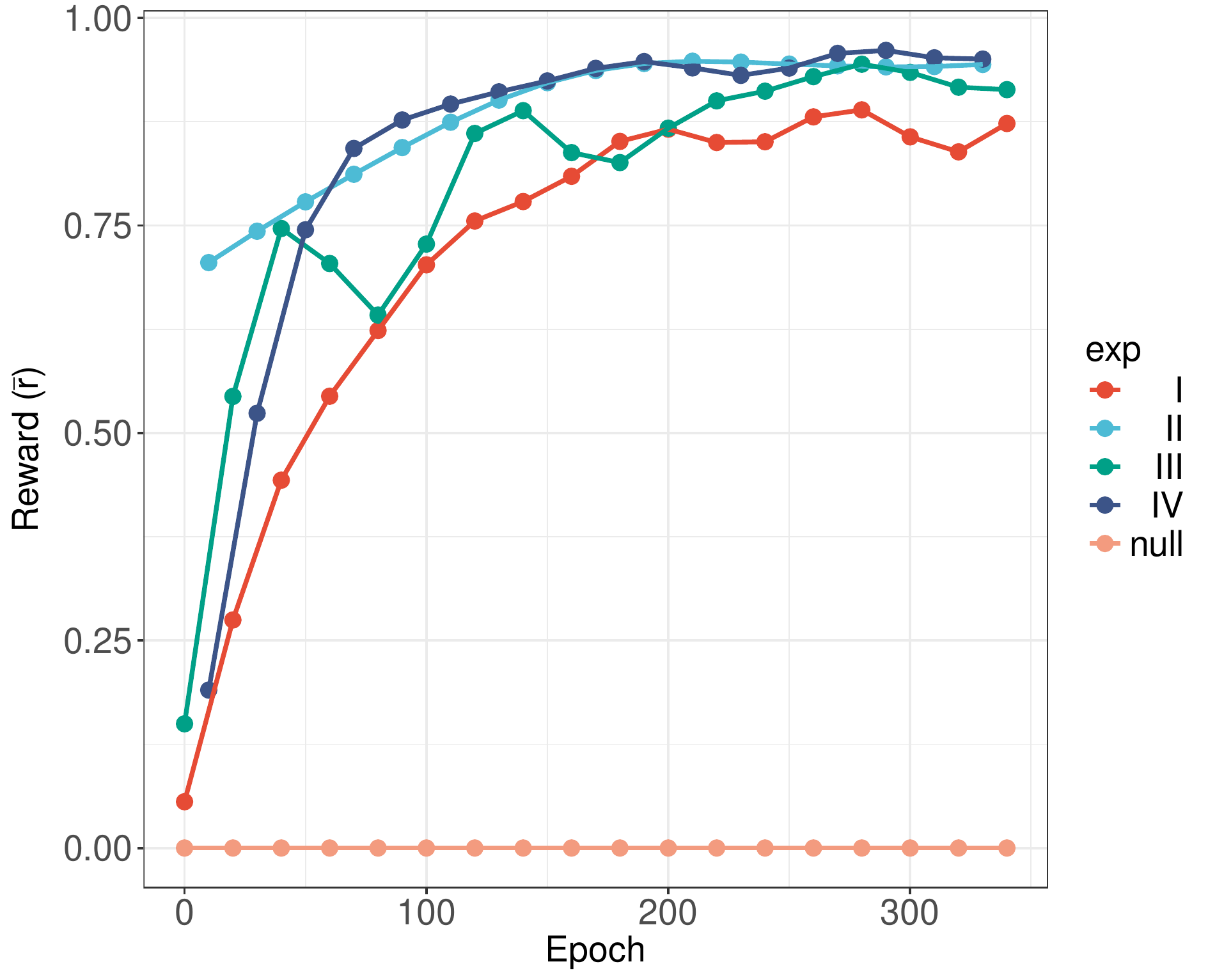}
\caption{Average reward}
\label{fig:experiment_total_reward}
\end{center}
\end{figure}

Figures \ref{fig:experiment_1_reward}, \ref{fig:experiment_1_fairnes} and \ref{fig:experiment_total_reward} show the evolution of the average cumulative bid, fairness and reward, respectively.
In experiment $\Romannum{1}$ \- the agent learns a policy that minimizes fairness average and maximizes bid average in contrast to experiment $\Romannum{2}$ in which fairness is maximized obtaining the minimum bid average. Note that the profit in experiment $\Romannum{1}$ is the highest because the algorithm learns to increase the bid value in groups $g_3, g_4$ and decrease the bid value in groups $g_1$ and $g_2$ (see table \ref{tab:segments_distributions} of $\bar{g}_\alpha$ values). In this policy, the bid values are adjusted to minimize the rejection ratio, defined as the price acceptance probability,  see equation (\ref{eq:elasticity}) and figure \ref{fig:elasticites}).

Experiments $\Romannum{3}$ and $\Romannum{4}$ are proposed with the aim of achieving a target fairness $f_t$. Indeed, the averaged values map to the selected targets given by $f_t$ ($f_t=0.90, \overline{f}=0.88$ and $f_t=0.75, \overline{f}=0.76$ for experiments $\Romannum{3}$ and $\Romannum{4}$, respectively). In experiment $\Romannum{3}$ the policy keeps the average price low on $g_3$ to reach a reasonable fairness ($\overline{f}=88\%$). Conversely in experiment $\Romannum{4}$ the calculated policy slightly increases the bid value in $g_3$ obtaining a less fair policy ($\overline{f}=76\%$) with more revenue. As a heavy penalization for bid rejections is used in the experiments, given by $\nu = -0.5$, the bid rejected ratio remains low. 

An interesting result arises from the comparison between fairness in experiments $\Romannum{2}$ and $\Romannum{5}$ (null model). The average prices in every group are $\sim 5.5$ in the null model because they are randomly and uniformly selected from $\mathcal{A}$. However, due to our definition of the rotated Jain's index in equation  (\ref{eq:jain_index}), the average fairness in $\Romannum{2}$ is slightly higher than that in  $\Romannum{5}$.


Related to the learning convergence, a descent in the average learning rate is a good indicator of learning convergence \cite{Watkins1992}. Let $p$ a pair with the form $(s,a \mid  s\in \mathcal{S}, a \in \mathcal{A})$ and $\mathcal{P}^n$ the set of all pairs $p$ visited by the agent. Then, the learning rate is defined as: 
\begin{equation}
\begin{split}
\label{eq:learning_rate}
l = \mathbb{E} \big[ \{ C(x) \mid \forall x\in\mathcal{P} \}\big], \text{ where } C(x) = \sum_{p \in \mathcal{P}} \delta_{x p}.
\end{split}
\end{equation}

\begin{figure}[h]
\begin{center}
\includegraphics[width=2.5in] {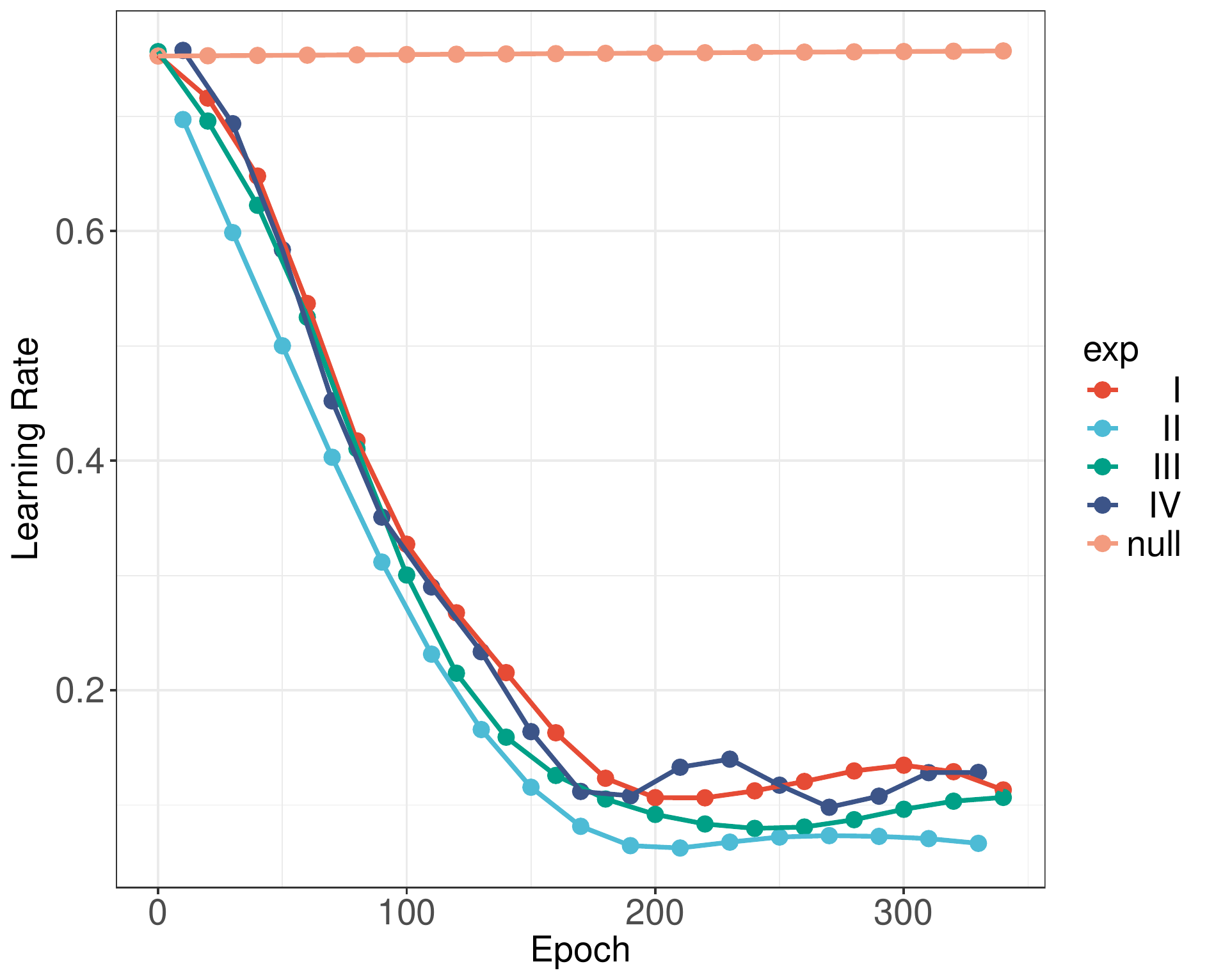}
\caption{Learning rate convergence}
\label{fig:learning_rate_convergence}
\end{center}
\end{figure}

Figure \ref{fig:learning_rate_convergence} shows the convergence rate of learning rates for the experiments proposed. As observed, the learning rate exhibits a descent to 0, guaranteeing the convergence of the algorithm.

\section{Conclusion}\label{secc:conclusions}

Fairness, as an ethical definition, is sometimes seen as negative by constraining developments \cite{bowieEthics}. However the present paper represents an example in which both customers and companies can benefit from transparency. Even if there is a disagreement regarding what is a fair policy in dynamic pricing, the fairness design principles proposed  in section \ref{secc:fairness_design} provide an initial definition based on equality, that constitutes a promising way to open a rich dialog. In this paper we have demonstrated that an unfair scenario can be \emph{unbiased} within a certain degree (the fairness targets given by $f_t$), by integrating fairness metrics as part of the model optimization. We cover the integration of fairness from a \emph{resource allocation point of view}, including it as a design principle. 

As future work, there are possible extensions to this study. Firstly, there are many parameter ($\lambda, \beta_p, \beta_f, \nu, ... $) to tuned in the learning process, considering the time required to evaluate each combination of parameters is a challenging task. Secondly, the comparison with other state-of-the-art models, like evolutionary algorithms \cite{evolutionaryAndRL}, is a valuable asset to an effective contrast to the Q-learning method proposed. Finally, more complex environment can be defined including more variables as time or external factors.

We demonstrate empirically that an effective balance between revenue and fairness maximization can be achieved in real time models like RL, in which the model learns by performing actions in an environment. We proposed a fairness metrics based on a rotated Jain's index. We have developed a synthetic environment to experiment with four different customer sensitivities. We have also tested several parametrized experiments, demonstrating that a given balance between revenue and fairness can be chosen.






%
\bibliographystyle{IEEEtran}

\end{document}